\documentclass[runningheads]{llncs}
\usepackage[T1]{fontenc}
\usepackage{graphicx}
\usepackage{algorithm}
\usepackage{algorithmic}
\usepackage{tikz}
\usetikzlibrary{arrows}

\begin{document}
\title{Nested Search versus Limited Discrepancy Search}
%
%
\author{Tristan Cazenave}
\authorrunning{Tristan Cazenave}
%
\institute{LAMSADE, Université Paris Dauphine - PSL, CNRS, Paris, France}
\maketitle              

\begin{abstract}
Limited Discrepancy Search (LDS) is a popular algorithm to search a state space with a heuristic to order the possible actions. Nested Search (NS) is another algorithm to search a state space with the same heuristic. NS spends more time on the move associated to the best heuristic playout while LDS spends more time on the best heuristic move. They both use similar times for the same level of search. We advocate in this paper that it is often better to follow the best heuristic playout as in NS than to follow the heuristic as in LDS.
\end{abstract}

\section{Introduction}

Combinatorial Optimization has many applications in numerous fields. Heuristic algorithms aim at solving Combinatorial Optimization problems using domain dependent knowledge without proving the optimality of the solution found. When the possible decisions can be sorted according to a heuristic, algorithms such as LDS \cite{Harvey1995} and NS \cite{tesauro1996line} can leverage the heuristic and find good solutions to difficult problems much faster than other more complete search algorithms.

Decision problems are a special kind of problems since finding a solution to a problem is equivalent to finding an optimal solution. Many Constraint Programming problems fall into this category. For this kind of problems LDS and NS can be particularly effective compared to complete search algorithms.

LDS is a popular search algorithm. There have been many papers using LDS to effectively search difficult problems including very recent ones. The base level performs playouts with a move ordering heuristic. A nice property of LDS is that it is guaranteed to improve on the previous levels with increasing levels of search. It has been applied to many difficult problems and it is currently commonly used for Combinatorial Optimization and Constraint Programming problems.

For example, LDS has been recently used as a search algorithm to combine Deep Reinforcement Learning and Constraint Programming in order to solve Combinatorial Optimization problems such as the Traveling Salesman Problem with Time Windows (TSPTW), 4-Moments Portfolio Optimization or the 0-1 Knapsack Problem \cite{cappart2021combining}.

It has also been recently used in combination with bandits in the Bandit Limited Discrepancy Search algorithm that was used for optimized algorithm selection in a fixed Machine Learning pipeline structure \cite{kishimoto2022bandit}.

NS is close to LDS but it differs in the choice of the decision that will be searched more. LDS follows the heuristic prior in order to decide which decision will be searched more. NS chooses to search more the decision that scores best according to the lower level playouts.

The paper is organized as follows. The second section recalls LDS. The third section deals with NS. The fourth section theoretically compares NS and LDS. The fifth section experimentally compares the two algorithms on four difficult Combinatorial Optimization problems.

\begin{algorithm}[tbh]
\begin{algorithmic}[1]
\STATE{LDS ($state$, $level$)}
\begin{ALC@g}
\IF{terminal ($state$)}
\RETURN{score ($state$)}
\ENDIF
\STATE{$best = -\infty$}
\STATE{sort the possible moves}
\FOR{$m$ in possible moves for $state$}
\STATE{$s$ $\leftarrow$ play ($state$, $m$)}
\IF{m is the first move}
\STATE{$score = $ LDS ($s$, $level$)}
\ELSIF{$level > 0$}
\STATE{$score = $ LDS ($s$, $level - 1$)}
\ELSE
\STATE{continue}
\ENDIF
\STATE{$best = max(best,score)$}
\ENDFOR
\RETURN{$best$}
\end{ALC@g}
\end{algorithmic}
\caption{\label{LDS}The LDS algorithm.}
\end{algorithm}

\tikzset{
  treenode/.style = {align=center, inner sep=0pt, text centered,
    font=\sffamily},
  arn_n/.style = {treenode, circle, white, font=\sffamily\bfseries, draw=black,
    text width=1.5em},
  arn_r/.style = {treenode, rectangle, draw=black,
    fill=black, minimum width=0.5em, minimum height=0.5em},
  arn_x/.style = {treenode, rectangle, draw=black,
    minimum width=0.5em, minimum height=0.5em},
  arn_e/.style = {treenode, circle, black, font=\sffamily\bfseries, 
  draw=black, text width=1.5em},
}

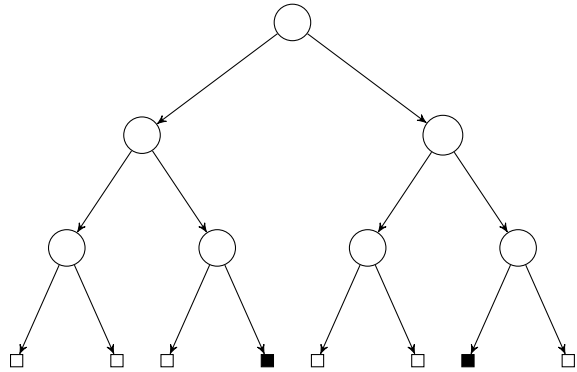
\begin{figure}
\begin{center}
\begin{tikzpicture}[->,>=stealth',level/.style={sibling distance = 4cm/#1,
  level distance = 1.5cm}] 
\node [arn_n] {}
    child{ node [arn_n] {} 
            child{ node [arn_n] {} 
            	child{ node [arn_x] {}}
				child{ node [arn_x] {}}
            }
            child{ node [arn_n] {}
							child{ node [arn_x] {}}
							child{ node [arn_r] {}}
            }                            
    }
    child{ node [arn_n] {47}
            child{ node [arn_n] {} 
							child{ node [arn_x] {}}
							child{ node [arn_x] {}}
            }
            child{ node [arn_n] {}
							child{ node [arn_r] {}}
							child{ node [arn_x] {}}
            }
		}
; 
\end{tikzpicture}
\caption{A binary search tree. The heuristic is to choose the left move. The black squares are the solutions. LDS does not find one of the two solutions using only one discrepancy since the path to the two solutions requires two discrepancies. A level 2 LDS solves the problem.}
\label{treeLDS}
\end{center}
\end{figure}

\section{Limited Discrepancy Search}

LDS \cite{Harvey1995} is given in Algorithm \ref{LDS}. It has applications in Constraint Programming and Combinatorial Optimization \cite{cappart2021combining}. It is a search strategy commonly used when a good heuristic on the possible actions is available for driving the search. The principle is to restrict the number of decisions deviating from the heuristic choices (i.e. a discrepancy) by a threshold. This will explore the subset of solutions that are likely to be good according to the heuristic, but it will also explore solutions where the heuristic has been reconsidered a given number of times (i.e. the number of discrepancies).

Figure \ref{treeLDS} gives an example of a binary search tree requiring a LDS of level 2 to be solved.

\begin{algorithm}[tbh]
\begin{algorithmic}[1]
\STATE{NS ($state$, $level$)}
\begin{ALC@g}
\IF{level == 0}
\RETURN{playout ($state$)}
\ENDIF
\WHILE{$state$ is not terminal}
\STATE{sort possible moves}
\FOR{$m$ in possible moves for $state$}
\STATE{$s \leftarrow$ play ($state$, $m$)}
\STATE{$s \leftarrow$ NS ($s$, $level-1$)}
\STATE{update $bestMove$ using score ($s$)}
\ENDFOR
\STATE{$state$ $\leftarrow$ play ($state$, $bestMove$)}
\ENDWHILE
\RETURN{$state$}
\end{ALC@g}
\end{algorithmic}
\caption{\label{NS}The NS algorithm.}
\end{algorithm}

\begin{algorithm}[tbh]
\begin{algorithmic}[1]
\STATE{playout ($state$)}
\begin{ALC@g}
\WHILE{$state$ is not terminal}
\STATE{$move$ $\leftarrow$ best heuristic move}
\STATE{$state$ $\leftarrow$ play ($state$, $move$)}
\ENDWHILE
\RETURN{$state$}
\end{ALC@g}
\end{algorithmic}
\caption{\label{playout}The playout algorithm}
\end{algorithm}

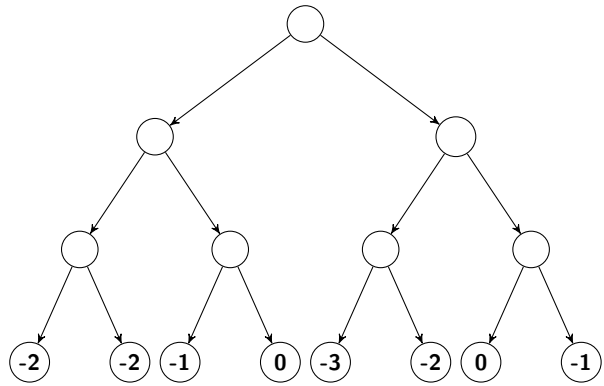
\begin{figure}
\begin{center}
\begin{tikzpicture}[->,>=stealth',level/.style={sibling distance = 4cm/#1,
  level distance = 1.5cm}] 
\node [arn_n] {}
    child{ node [arn_n] {} 
            child{ node [arn_n] {} 
            	child{ node [arn_e] {-2}}
				child{ node [arn_e] {-2}}
            }
            child{ node [arn_n] {}
							child{ node [arn_e] {-1}}
							child{ node [arn_e] {0}}
            }                            
    }
    child{ node [arn_n] {47}
            child{ node [arn_n] {} 
							child{ node [arn_e] {-3}}
							child{ node [arn_e] {-2}}
            }
            child{ node [arn_n] {}
							child{ node [arn_e] {0}}
							child{ node [arn_e] {-1}}
            }
		}
; 
\end{tikzpicture}
\caption{The same tree as in Figure \ref{treeLDS} with scores of the playouts at the leaves. A score of 0 means the problem is solved. A NS of level 1 finds a solution as it uses the scores of the playouts to direct its search.}
\label{treeNS}
\end{center}
\end{figure}

\section{Nested Search}

\subsection{Related Work}

NS has its root in Monte Carlo Search \cite{tesauro1996line}. An application then was then a Backgammon program which was improved thanks to nested rollouts. It was subsequently used in Combinatorial Optimization \cite{bertsekas1997rollout} and in stochastic scheduling problems \cite{bertsekas1999rollout}. Nested rollouts combined with a heuristic to choose the next move at the base level were used to improve a Klondike solitaire program \cite{yan2004solitaire}. Nested rollouts have been used with heuristics that change with the stage of the game of Thoughtful Solitaire, a version of Klondike Solitaire in which the locations of all cards is known \cite{bjarnason2007searching}. They were also used to learn a control policy for planning \cite{fern2004learning}.

Nested Monte Carlo Search (NMCS) \cite{CazenaveIJCAI09} is
a related algorithm that works well for puzzles and optimization problems. It biases its playouts using lower level playouts. At level zero NMCS adopts a randomized playout policy. The main improvement of NMCS over NS is the memorization of the best sequence at each recursion level. Without the memorization of the best sequence the algorithm does not give good results for levels greater than one. Applications of NMCS include Single Player General Game Playing \cite{Mehat2010}, Cooperative Pathfinding \cite{Bouzy13}, Software testing \cite{PouldingF14}, heuristic Model-Checking \cite{PouldingF15}, the Pancake problem \cite{Bouzy16}, Games \cite{CazenaveSST16} and the RNA inverse folding problem \cite{portela2018unexpectedly}. Online learning of playout strategies combined with NMCS has given good results on optimization problems \cite{RimmelEvo11}. Online learning of a playout policy in the context of nested searches has been further developed for puzzles and optimization with Nested Rollout Policy Adaptation (NRPA) \cite{Rosin2011}. NRPA has found new world records in Morpion Solitaire and crosswords puzzles. Edelkamp, Cazenave and co-workers have applied the NRPA algorithm to multiple problems: The TSPTW problem \cite{cazenave2012tsptw,edelkamp2013algorithm}, 3D Packing with Object Orientation \cite{edelkamp2014monte}, the physical traveling salesman problem \cite{edelkamp2014solving}, the Multiple Sequence Alignment problem \cite{edelkamp2015monte}, Logistics \cite{edelkamp2016monte,Cazenave2021Policy}, Graph Coloring \cite{Cazenave2020Graph} and Inverse Folding \cite{Cazenave2020Inverse}. The principle of NRPA is to adapt the playout policy reinforcing the moves of the best sequence of moves found so far at each level.

\subsection{The Algorithm}

At the lowest recursive level of NS, the generation of sequences is driven by the heuristic on decisions. Sequences are generated with the heuristic as described in Algorithm~\ref{playout}.  

In NS, the base heuristic remains the same throughout the execution of the algorithm. However, the heuristic is combined with a tree search to improve the quality over a simple sequence generator. At each step, each possible move is evaluated by completing the partial solution into a complete one using moves sampled from the underlying recursion level. Whichever intermediate move has led to the best completed sequence, is selected and added to the current sequence. The same procedure is repeated to choose the following move, until the sequence has reached a terminal state. See Algorithm~\ref{NS}.

Figure \ref{treeNS} gives an example of a binary search tree where the scores of the playouts enable a level 1 NS to solve the problem.

\section{Comparison of NS and LDS}

\subsection{Complexity}

For a tree of height h and branching factor b, the total number of playout steps of a NS of level n will be $t_n(h, b) =  t_n(h-1,b) + b \times t_{n-1}(h-1,b)$. The total number of playout steps of a LDS of level n will be $t_n(h, b) = t_{n}(h-1,b) + (b - 1) \times t_{n-1}(h-1,b)$. The difference between the complexity of two searches is $t_{n-1}(h-1,b)$. The complexity of a NS of level n is O($b^n \times h^{n+1}$) \cite{Mehat2010}. So $t_{n-1}(h-1,b)$ is small compared to $t_n(h-1, b)$ and therefore NS and LDS have similar complexities even if LDS can be slightly faster. However if the complexity is also related to a better move ordering such as in decision problems, NS with high levels can become faster than LDS with high levels since NS spends much more resources on the choice of the move to search more. 

\subsection{Analysis}

The main difference between NS and LDS is the exploitation of the move associated to the best playout of the lower level as in NS instead of the exploitation of the move advised by the static heuristic as in LDS. In NS the playouts are played with the heuristic as in LDS so NS also benefits from the heuristic. Moreover, NS uses the scores of playouts.

NS uses the scores at the end of the playouts to exploit the most promising move. On the contrary LDS does not require a score for playouts but only for possible moves. For many problems the score of a playout is natural as it is the value to optimize. For constraint satisfaction problems however the score has to be designed. A possible scoring function for NS is to count the number of unassigned variables when reaching an inconsistency and to return the opposite of the count as NS maximizes the score.

The order in which LDS and NS search the state space is quite different. LDS starts by going very deep in the part of the search tree where the heuristic on moves leads. It then backtracks exploring with the maximum exploration the bottom of this part of the tree. NS has a different behavior. It goes in the bottom of the tree with playouts and carefully chooses the move that it plays in the top of the tree before exploring it more. 

So LDS backtracks from the bottom while NS searches with decreasing levels at the top of the search tree. When the heuristic does not fail at the top of the search tree LDS can be faster since it backtracks first in the most interesting parts of the search tree. However for difficult problems, high levels of search are often required and the heuristic is more difficult to design near the root than near the leaves. So NS which more carefully chooses its moves near the root can prove better.

\section{Experimental Results}

In this section we present experiments comparing NS and LDS for four difficult combinatorial problems. All experiments were done on AMD EPYC-Rome 2 GHz processors with 512 KB of cache.

\subsection{TSPTW}

The TSPTW is a practical problem that has everyday applications. It has time constraints represented as time intervals during which cities are to be visited. In LDS and NS paths with violated constraints can be generated. As presented in \cite{RimmelEvo11} , a new score $Tcost(p)$ of a path $p$ can be defined as follow:
$$
Tcost(p) = -cost(p) - 10^6 * \Omega(p),
$$
with, $cost(p)$ the sum of the distances of the path $p$ and $\Omega(p)$ the number of violated constraints. $10^6$ is a constant chosen high enough so that the algorithm first optimizes the constraints then the sum of the distances.

We experiment with NS and LDS for the TSPTW problem with most of the problems of the standard test set \cite{potvin1996vehicle}. Problems that took a very long time for LDS(3) and NS(3) were removed. The results are given in Table \ref{tsptw}. We can observe that for level 1, LDS is strictly better than NS 8 times while NS is strictly better than LDS 15 times. At level 3 NS becomes strictly better 17 times while LDS is strictly better only twice. When the level increases NS makes a better use of the lower level playouts to direct its choices than LDS since LDS still uses the heuristic to order moves and to choose the best move that will benefit from the intensification of the search.

\begin{table*}[!t]
\centering
\begin{tabular}{l | r r | r r | r r }
Instance  & LDS (1) & Nested (1) & LDS (2) & Nested (2) & LDS (3) & Nested (3) \\ \hline
rc\_201.1 & \textbf{-1000451.06} & -1000461.69 & \textbf{-445.38} & -447.29 & \textbf{-444.54} & \textbf{-444.54}\\
rc\_201.4 & -4000771.50 & \textbf{-1000796.44} & -2000781.75 & \textbf{-1000770.19} & -1000778.44 & \textbf{-793.64}\\
rc\_202.3 & -10000946.00 & \textbf{-4000886.00} & \textbf{-4000862.25} & -4000878.75 & -2000853.50 & \textbf{-859.53}\\
rc\_203.2 & \textbf{-3000892.25} & -4000880.00 & -1000833.31 & \textbf{-948.24} & -854.54 & \textbf{-850.68}\\
rc\_205.1 & \textbf{-353.17} & -1000370.56 & \textbf{-343.36} & -379.48 & \textbf{-343.21} & \textbf{-343.21}\\
rc\_205.4 & -13000789.00 & \textbf{-9000761.00} & -9000741.00 & \textbf{-1000808.56} & -3000775.00 & \textbf{-1000757.69}\\
rc\_206.3 & -15000839.00 & \textbf{-7000676.00} & -3000656.75 & \textbf{-597.70} & -1000609.25 & \textbf{-592.83}\\
rc\_207.2 & \textbf{-17000822.00} & -25000838.00 & -13000822.00 & \textbf{-12000796.00} & -11000883.00 & \textbf{-827.96}\\
rc\_201.2 & -18000820.00 & \textbf{-6000777.50} & -9000754.00 & \textbf{-733.85} & -1000738.00 & \textbf{-733.85}\\
rc\_202.1 & -8000814.00 & \textbf{-6000811.00} & -5000802.00 & \textbf{-3000849.00} & -3000759.00 & \textbf{-802.87}\\
rc\_202.4 & -8000809.50 & \textbf{-2000818.62} & -5000866.50 & \textbf{-846.50} & -3000858.00 & \textbf{-816.09}\\
rc\_203.3 & -8000855.00 & \textbf{-4000920.50} & -5000861.00 & \textbf{-2000884.25} & -3000865.50 & \textbf{-848.09}\\
rc\_205.2 & -18000944.00 & \textbf{-10000778.00} & -6000756.00 & \textbf{-1000798.56} & -2000737.75 & \textbf{-755.93}\\
rc\_206.1 & \textbf{-117.85} & \textbf{-117.85} & \textbf{-117.85} & \textbf{-117.85} & \textbf{-117.85} & \textbf{-117.85}\\
rc\_206.4 & -16000943.00 & \textbf{-10000930.00} & -13000922.00 & \textbf{-6001000.00} & -10000873.00 & \textbf{-889.23}\\
rc\_207.3 & -20000940.00 & \textbf{-15000885.00} & -12000857.00 & \textbf{-818.64} & -6000798.00 & \textbf{-713.89}\\
rc\_201.3 & -20000848.00 & \textbf{-13000891.00} & -14000837.00 & \textbf{-7000819.00} & -7000823.50 & \textbf{-6000851.50}\\
rc\_202.2 & \textbf{-309.94} & -310.59 & -305.76 & \textbf{-304.14} & \textbf{-304.14} & \textbf{-304.14}\\
rc\_203.1 & \textbf{-1000463.44} & \textbf{-1000463.44} & \textbf{-461.39} & -467.69 & \textbf{-458.93} & -459.54\\
rc\_203.4 & -316.19 & \textbf{-314.29} & \textbf{-314.29} & \textbf{-314.29} & \textbf{-314.29} & \textbf{-314.29}\\
rc\_204.3 & \textbf{-486.88} & -1000495.88 & \textbf{-463.43} & -473.12 & \textbf{-459.03} & -461.93\\
rc\_205.3 & -5000917.50 & \textbf{-2000835.88} & -3000832.25 & \textbf{-2000822.75} & -2000837.25 & \textbf{-867.33}\\
rc\_206.2 & -21001112.00 & \textbf{-16000962.00} & -15000968.00 & \textbf{-10000882.00} & -5000882.00 & \textbf{-1000871.25}\\
rc\_207.1 & \textbf{-7000871.00} & -8000913.50 & -2000790.00 & \textbf{-849.65} & -1000767.50 & \textbf{-781.54}\\
rc\_207.4 & \textbf{-119.64} & \textbf{-119.64} & \textbf{-119.64} & \textbf{-119.64} & \textbf{-119.64} & \textbf{-119.64}\\ \hline
Better & 8 & \textbf{15} & 5 & \textbf{17} & 2 & \textbf{17}\\
\end{tabular}
\caption{Results for the TSPTW problem instances}
\label{tsptw}
\end{table*}

\subsection{RNA Inverse Folding}

The design of molecules with specific properties is an important topic for health related research. RNA molecules are long molecules composed of four possible nucleotides. Molecules can be represented as strings composed of the four characters A, C, G, U. For RNA molecules of length N, the size of the state space of possible strings is exponential in N. It can be very large for long molecules. The sequence of nucleotides folds back on itself to form what is called its secondary structure. It is possible to find in a polynomial time the folded structure of a given sequence. However, the opposite, which is the RNA inverse folding problem, is much harder and is supposed to be NP-complete. 

\begin{figure}
    \centering
    \includegraphics[width=6cm]{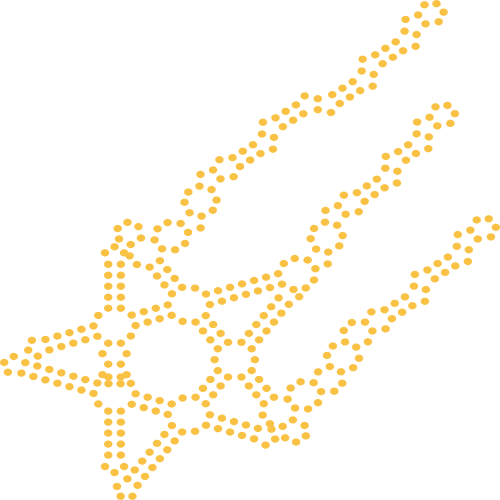}
    \caption{Example of a RNA design puzzle from Eterna.}
    \label{star}
\end{figure}

We compare LDS and NS on the Eterna100 benchmark which contains 100 RNA secondary structure puzzles of varying degrees of difficulty. A puzzle consists of a given structure under the dot-bracket notation. This notation defines a structure as a sequence of parentheses and points each representing a base. The matching parentheses symbolize the paired bases and the dots the unpaired ones. The puzzle is solved when a sequence of the four nucleotides A,U,G and C, folding according to the target structure, is found. In some puzzles, the value of certain bases is imposed. Figure \ref{star} gives an example of an Eterna100 problem.

Where human experts have managed to solve the 100 problems of the benchmark, no program has so far achieved such a score. The best score so far is 95/100 by NEMO, NEsted MOnte Carlo RNA Puzzle Solver \cite{portela2018unexpectedly} and by GNRPA \cite{Cazenave2020Inverse}.

We use a the NEMO heuristic to order moves for LDS and NS. It mainly consists in probabilities for pair of bases.

Table \ref{RNA} gives the number of problems solved by LDS and NS for levels 1 to 3. NS is much better than LDS for all levels.

Since NS solves RNA inverse folding problems faster than LDS and since inverse folding is a decision problem, the total execution time for all the problems of NS is smaller than that of LDS.

\begin{table*}[!t]
\centering
\begin{tabular}{r r | r r | r r }
LDS (1) & NS (1) & LDS (2) & NS (2) & LDS (3) & NS (3) \\ 
24 & \textbf{32} & 30 & \textbf{59} & 39 & \textbf{75}\\
\end{tabular}
\caption{Number of problems solved for the Eterna100 benchmark.}
\label{RNA}
\end{table*}

\subsection{The Snake-in-the-Box}

The Snake-in-the-Box is a problem from graph theory with applications in coding theory, disjunctive normal form simplification, electrical engineering, analog-to-digital conversion, electronic combination locking, and computer network topologies \cite{krafka2010snake}.

The problem is to find the longest possible path in a N-dimensional hypercube so that any vertex has at most two other neighboring vertices that are also in the graph. For example in dimension 3, the graph is a cube with 8 vertices: 000, 001, 010, 011, 100, 101, 110 and 111. The longest path is 000, 001, 011, 111, 110. It has a length of 4, so the best score for dimension 3 is 4. For dimensions greater than 9 the exact scores are still unknown and only lower bounds are known.

Kinny broke world records at the Snake-in-the-Box applying Nested Monte Carlo Search \cite{Kinny2012}. He used a heuristic to order move in the playouts. The heuristic is to favor moves that lead to a state where there is only one possible move. Nested Rollout Policy Adaptation combined to beam search also reached high scores at the Snake-in-the-Box \cite{Edelkamp16Diversity}.

In our experiments we use a heuristic similar to the heuristic used by Kinny. We sort the moves according to the number of possible moves after the move. If there are no possible moves the move is associated to a large penalty, otherwise the score of a move is the number of possible moves after the move. Moves are sorted by ascending scores.

Table \ref{Snake} give the results for LDS and NS of levels 1 to 3. We can observe that for simple problems LDS of level 1 is better than NS of level 1 but for dimensions 10 and 11 that are harder NS gets better. At level 3 NS is always better than LDS. This is probably due to the fact that at the top level NS of level 3 uses level 2 playouts to order moves when LDS still uses the heuristic.

\begin{table*}[!t]
\centering
\begin{tabular}{l | r r | r r | r r }
Dimension  & LDS (1) & NS (1) & LDS (2) & NS (2) & LDS (3) & NS (3) \\ \hline
8 & \textbf{84.00} & 79.00 & \textbf{85.00} & 84.00 & 89.00 & \textbf{91.00}\\
9 & \textbf{149.00} & 148.00 & 159.00 & \textbf{166.00} & 161.00 & \textbf{167.00}\\
10 & 266.00 & \textbf{271.00} & 283.00 & \textbf{298.00} & 301.00 & \textbf{310.00}\\
11 & 467.00 & \textbf{493.00} & 515.00 & \textbf{543.00} & 532.00 & \textbf{570.00}\\
\end{tabular}
\caption{Results for the Snake-in-the-Box.}
\label{Snake}
\end{table*}

\begin{table*}
\centering
\begin{tabular}{ r r r r r r r r r r r }
0,0 & 1,2 & 2,5 & 3,9 & 4,8 & 5,4 & 6,3 & 7,1 & 8,10 & 9,7 & 10,6\\
1,1 & 2,3 & 3,6 & 4,10 & 5,9 & 6,5 & 7,4 & 8,2 & 9,0 & 10,8 & 0,7\\
2,2 & 3,4 & 4,7 & 5,0 & 6,10 & 7,6 & 8,5 & 9,3 & 10,1 & 0,9 & 1,8\\
3,3 & 4,5 & 5,8 & 6,1 & 7,0 & 8,7 & 9,6 & 10,4 & 0,2 & 1,10 & 2,9\\
4,4 & 5,6 & 6,9 & 7,2 & 8,1 & 9,8 & 10,7 & 0,5 & 1,3 & 2,0 & 3,10\\
5,5 & 6,7 & 7,10 & 8,3 & 9,2 & 10,9 & 0,8 & 1,6 & 2,4 & 3,1 & 4,0\\
6,6 & 7,8 & 8,0 & 9,4 & 10,3 & 0,10 & 1,9 & 2,7 & 3,5 & 4,2 & 5,1\\
7,7 & 8,9 & 9,1 & 10,5 & 0,4 & 1,0 & 2,10 & 3,8 & 4,6 & 5,3 & 6,2\\
8,8 & 9,10 & 10,2 & 0,6 & 1,5 & 2,1 & 3,0 & 4,9 & 5,7 & 6,4 & 7,3\\
9,9 & 10,0 & 0,3 & 1,7 & 2,6 & 3,2 & 4,1 & 5,10 & 6,8 & 7,5 & 8,4\\
10,10 & 0,1 & 1,4 & 2,8 & 3,7 & 4,3 & 5,2 & 6,0 & 7,9 & 8,6 & 9,5\\
\end{tabular}
\label{Square11}
\caption{Example of a Graeco-Latin square of size 11 found by LDS in 0.00 seconds.}
\end{table*}

\begin{table}
\centering
\begin{tabular}{ r r r r r r r r r r r }
0,0 & 1,1 & 2,2\\
1,2 & 2,0 & 0,1\\
2,1 & 0,2 & 1,0\\
\end{tabular}
\caption{Example of a Graeco-Latin square of size 3.}
\label{Square3}
\end{table}

\subsection{Graeco-Latin Squares}

A Latin square is a $n \times n$ matrix containing $n$ symbols such that each symbol occurs only once in each row and each column. A Graeco-Latin square is the association of two Latin squares of the same dimension such that when they are superimposed every pair of symbols occurs only once. Table \ref{Square3} gives an example of a small Graeco-Latin square.

Graeco-Latin squares have a long history \cite{styan2009some}. They are also called Euler squares since Euler proposed a famous conjecture about these squares. Euler first considered the problem of the 36 officers \cite{euler1782recherches}. This problem is equivalent to constructing a $6 \times 6$ Graeco-Latin square. Euler conjectured in 1779 it is not possible, and Tarry \cite{tarry1900} proved it in 1900. Euler also conjectured that there are no Graeco-Latin squares of order 10, 14, 18, and so on. This conjecture was proved false in 1959 \cite{bose1959falsity}.

Graeco-Latin squares are used for designing experiments in biology, medicine, sociology and even marketing \cite{gardner1995euler}. Graeco-Latin squares have even been used in literature: Georges Perec used a $10 \times 10$ Graeco-Latin square as a basis for his novel "La Vie mode d’emploi" \cite{perec1978vie}.

In order to model the problem for LDS and NS we maintain the possible values of each cell after each move. A move is the assignment of a value to a variable. We remove the assigned value from the domains of the variables in the same row and in the same column as the variable being assigned. We also remove the value from variables in the same square that are associated in the other square to the same value as the value associated to the variable being assigned. We also perform the symmetrical removal of the values in the other square that are associated in the current square to the value being assigned.

We tested two variable ordering heuristics. The dom heuristic that was found effective for Latin Squares \cite{dotu2003redundant}. It selects the variable with the smallest number of possible values. The deg heuristic that selects the variable that is connected to the least number of other free variables in the same row and the same column. The deg heuristic keeps more possible values overall but it also has a greater branching factor than dom.

A possible move is an assignment of a value to the chosen variable. The list of possible moves is the list of values that can be assigned to the chosen variable. Move ordering is done sorting the moves from the moves that remove the smallest number of possible values from other variables to the moves that removes the greatest number of possible values from other variables.

We also use an algorithm we call MAC, since it is related to arc consistency, to prune more values and to detect inconsistencies earlier. The algorithm uses channeling constraints as for Latin Squares \cite{dotu2003redundant}. The algorithm stops the search if a variable has an empty domain and recursively assign variables that only have one possible value. It also calculates for each color and each row the number of times the color is present in the row and assigns the color if it is present only once or stops the search if it is not present. It does the same for the columns. This algorithm is usually effective for solving Latin Squares since it reduces a lot the search space.

A state is terminal either if all variables have been assigned or if a non assigned variable has an empty domain. The score function is simply the opposite of the number of free variables.

We also tested symmetry breaking  as described in \cite{appa2006searching,rubin2021improving}. To do so, we fix the first row of every square to be in lexicographic order which eliminates the permutations of the columns and fixes the symbol permutations to be the same in each square. We also fix the first column of the first square to be in lexicographic order which eliminates permutations of the rows and perform domain reduction of
entries in the first column of the second square.

Table \ref{GraecoLatin} compares LDS and NS for different dimensions, levels and parameters of the search algorithms. The timeout is fixed at 10 000 seconds. The var column indicates the heuristic used to select the next variable to assign, the sym column indicates if symmetry breaking is used or not, the MAC column indicates if arc-consistency is used or not. We ran the different combination of the algorithms for levels 3 to 6 and dimensions 5 to 11 (except 6 since it is not solvable). If a time is given it means the problem was solved within this time. If a score and a time are given it means the problem was not solved, the score is the score resulting from the search (the opposite of the number of free variables) and the time is the time taken for the search. A bold cell mean this is the best result for this dimension.

Overall NS and LDS solve immediately all dimensions, except dimension 10, when called with the appropriate parameters. It is noticeable that in previous work some dimensions took much more time to be solved. For example a SAT solver take 45 minutes to solve size 9 \cite{bright2019effective}. Integer Programming with the Gurobi solver, which is the state of the art solver, solves size 9 in 344 seconds \cite{rubin2021improving} and Constraint Programming takes from 1 552 to 50 753 seconds for size 9 \cite{rubin2021improving}. However for the more difficult problem of size 10 we do not find a solution within 10 000 seconds when SAT finds it in 23 hours, Integer Programming finds it in 3 046 seconds and Constraint Programming finds it in 11 971 seconds.

We can also observe that the deg variable ordering heuristic is much better than the dom variable ordering heuristic in all cases. The principle here is to constrain less the problem and let as much freedom as possible to the algorithm, reducing less the number of possible values. This is interesting since when the goal is to prove that a state is not possible to solve it is better to constrain as much as possible the possible value in order to save search time. Here the principle when looking for a solution is the opposite: let as many options as possible. This opposition of principles can also be seen for the symmetry breaking option. Breaking symmetries reduces the options which is good for a complete search but which is here harmful when looking for an unique assignment. We also see for dimension 11 that NS solves immediately the problem without MAC but fails to solve it with MAC. This is another illustration of this principle. However MAC can be useful as can be seen from dimension 8 where the best results for NS and LDS were obtained with MAC.

\begin{table*}
\centering
\begin{tabular}{l l l l l r r r r}
Dimension  & Algorithm & var & sym & MAC & level 3 & level 4 & level 5 & level 6\\[0.5ex]
\hline\\ [-1.5ex]
5          &    NS & deg & yes & yes & \textbf{0.00s} & \textbf{0.00s} & \textbf{0.00s} & \textbf{0.00s}\\
5          &       LDS & deg & yes & yes & \textbf{0.00s} & \textbf{0.00s} & \textbf{0.00s} & \textbf{0.00s}\\ [0.5ex]
\hline\\ [-1.5ex]
7          &    NS & deg & yes & yes & -5 in 0.91s & 0.76s & 0.94s & 1.10s\\
7          &    NS & dom & yes & yes & -9 in 1.52s & 6.55s & 14.51s & 81.93s\\
7          &    NS & deg & no & yes & \textbf{0.00s} & \textbf{0.00s} & \textbf{0.00s} & \textbf{0.00s}\\
7          &    NS & deg & no & no & \textbf{0.00s} & \textbf{0.00s} & \textbf{0.00s} & \textbf{0.00s}\\
7          &       LDS & deg & yes & yes & 0.32s & 0.80s & 1.03s & 2.63s\\
7          &       LDS & dom & yes & yes & -13 in 0.50s & -9 in 2.62s & 11.60s & 4.23s\\
7          &       LDS & deg & no & yes & \textbf{0.00s} & \textbf{0.00s} & \textbf{0.00s} & \textbf{0.00s}\\
7          &       LDS & deg & no & no & \textbf{0.00s} & \textbf{0.00s} & \textbf{0.00s} & \textbf{0.00s}\\[0.5ex]
\hline\\ [-1.5ex]
8          &    NS & deg & yes & yes & -7 in 12.62s & -5 in 242.95s & 3172.59s & 1493.77s\\
8          &    NS & dom & yes & yes & -14 in 13.71s & -10 in 275.11s & -9 in 4360.04s & Timeout\\
8          &    NS & deg & no & yes & -7 in 177.00s & \textbf{5.45s} & 277.91s & 9609.19s\\
8          &    NS & deg & no & no & -14 in 84.77s & -10 in 4041.90s & 3553.67s & 2992.09s\\
8          &       LDS & deg & yes & yes & -8 in 6.00s & -6 in 86.46s & -5 in 1000.00s & 4311.28s \\
8          &       LDS & dom & yes & yes & -16 in 4.90s & -13 in 61.31s & 439.10s & 3446.18s\\
8          &       LDS & deg & no & yes & 63.03s & 46.84s & 512.79s & 4561.50s\\
8          &       LDS & deg & no & no & -14 in 22.74s & 239.25s & 272.16s & 2531.84s\\[0.5ex]
\hline\\ [-1.5ex]
9          &    NS & deg & yes & yes & -10 in 100.59s & -9 in 2745.83s & Timeout & Timeout\\
9          &    NS & dom & yes & yes & -19 in 105.17s & -15 in 3149.61s & Timeout & Timeout\\
9          &    NS & deg & no & yes & \textbf{0.00s} & \textbf{0.00s} & \textbf{0.00s} & \textbf{0.00s}\\
9          &    NS & deg & no & no & \textbf{0.00s} & \textbf{0.00s} & \textbf{0.00s} & \textbf{0.00s}\\
9          &       LDS & deg & yes & yes & 0.03s & 0.09s & 0.18s & 0.13s\\
9          &       LDS & dom & yes & yes & -22 in 38.04s & -15 in 831.53s & Timeout & Timeout\\
9          &       LDS & deg & no & yes & \textbf{0.00s} & \textbf{0.00s} & \textbf{0.00s} & \textbf{0.00s}\\
9          &       LDS & deg & no & no &  \textbf{0.00s} & \textbf{0.00s} & \textbf{0.00s} & \textbf{0.00s}\\[0.5ex]
\hline\\ [-1.5ex]
10          &    NS & deg & yes & yes & -12 in 557.21s & Timeout & Timeout & Timeout\\
10          &    NS & dom & yes & yes & -24 in 578.75 & Timeout & Timeout & Timeout\\
10          &    NS & deg & no & yes & -9 in 4036.96s & Timeout & Timeout & Timeout\\
10          &    NS & deg & no & no & -16 in 1626.02s & Timeout & Timeout & Timeout\\
10          &       LDS & deg & yes & yes &  -14 in 281.58s & Timeout & Timeout & Timeout\\
10          &       LDS & dom & yes & yes & -25 in 227.21s & -21 in 7892.41s & Timeout & Timeout\\
10          &       LDS & deg & no & yes & \textbf{-7 in 2315.48s} & Timeout & Timeout & Timeout\\
10          &       LDS & deg & no & no & -19 in 419.67s & Timeout & Timeout & Timeout\\[0.5ex]
\hline\\ [-1.5ex]
11          &    NS & deg & yes & yes & -15 in 2504.20s & Timeout & Timeout & Timeout\\
11          &    NS & dom & yes & yes & -33 in 2653.31s & Timeout & Timeout & Timeout\\
11          &    NS & deg & no & yes & Timeout & Timeout & Timeout & Timeout \\
11          &    NS & deg & no & no & \textbf{0.00s} & \textbf{0.00s} & \textbf{0.00s} & \textbf{0.00s}\\
11          &       LDS & deg & yes & yes & -12 in 1376.73s & 39.68s & 17.83s & 61.69s\\
11          &       LDS & dom & yes & yes & -36 in 1050.09s & Timeout & Timeout & Timeout\\
11          &       LDS & deg & no & yes & 0.01s & \textbf{0.00s} & \textbf{0.00s} & \textbf{0.00s}\\
11          &       LDS & deg & no & no & \textbf{0.00s} & \textbf{0.00s} & \textbf{0.00s} & \textbf{0.00s}\\
\end{tabular}
\caption{Results for the Graeco-Latin squares.}
\label{GraecoLatin}
\end{table*}

\section{Conclusion}

We have theoretically compared NS and LDS, finding the two search algorithms have similar complexities. We also have analyzed them, conjecturing that for high levels of search NS can have a better behavior than LDS due to a better choice of the move to intensify at every node.

Experimental results on four difficult combinatorial optimization problems that have quite different properties confirm that for high levels of search NS often performs better than LDS.



\bibliographystyle{splncs04}
\bibliography{main}

\end{document}